# Artificial Learning in Artificial Memories

John Robert Burger
Professor Emeritus
Department of Electrical and Computer Engineering
25686 Dahlin Road
Veneta, OR 97487
(jrburger1@gmail.com)

*Abstract* – Memory refinements are designed below to detect those sequences of actions that have been repeated a given number n.  Subsequently such sequences are permitted to run without CPU involvement.  This mimics human learning.  Actions are rehearsed and once learned, they are performed automatically without conscious involvement.

**Introduction**

This paper is written from the perspective of designing artificial brains that are modeled after human brains.  This architectural work pays attention to human psychology including the interplay between short and long term memory, the realization of which is assumed constrained by circuit and system theory.  This paper is a humble attempt, but at least it has an original aspect.  My fellow engineering stereotypes tend to ignore human psychology; worse yet, scientists in life sciences, seem unaware of the constraints imposed by working circuits and systems.  Although their products such as long term potentiation are impressive at first sight and sometimes work in cyberspace, they are most unpromising for practical humanoid realizations.

Learning is much more than random or mathematical optimization leading to a successful sequence of actions, although no doubt successful actions need to be discovered.  The discovery of successful actions ought to be termed *searching* or *investigating*, anything but learning.  Learning is a specific attribute beyond mere memory that implies a long term physical modification of underlying circuitry such that, in the big picture, human efficacy increases.  Computers today obviously cannot be modified in any fundamental way by their own programming.  Consequently, the envisioned learning, as below, is unrelated to the field of *machine learning*.  Machine learning is traditionally limited to the programming of structured, fundamentally unalterable computers (Sun & Giles, 2001; Alpaydin 2010).

There are times when we might benefit from artificial learning in everyday life.  For example, a person might perform certain keystrokes over and over; one opens a mail server, the inbox is opened, email is selected in chronological order and it is properly filed, each action hundreds of times per year.  Yet today's best personal computers will not learn even this!

This paper is not about the programming of artificial neural networks (Haykin 1994).  The training of artificial neural networks using, for example, gradient descent, has been construed as learning, but properly this is not learning at all.  It is an exercise in mathematical optimization without biological underpinning.

**Learning Modeled After Biology**
Learning may be viewed in the biological sense as an internalization process that detects, remembers and reproduces successful sequences.  Learning results in performing an action, or possibly several actions concurrently without cognitive effort or thinking, meaning without conscious involvement.

Memory hardware is designed below to learn repeated actions, analogous to rehearsal when humans learn.  Much of what we learn is a sequence of actions, for examples, proper components of walking, reciting a poem or reproducing a musical tune.  Such things are memory-mapped actions and may be performed without CPU effort (analogous to thinking) once they have been learned, and this is a key point.  Properties of artificial learning for memory-mapped sequences are summarized in Table 1.

**Table 1   Artificial Learning**

| 1 | A result of rehearsal |
|---|---|
| 2 | Occurs within long term memory |
| 3 | Enables long term modification of underlying circuitry |
| 4 | Permits action without CPU effort |
| 5 | Permits action without CPU delays |
| 6 | Permits action without CPU memory usage |
| 7 | Independent sequences may run concurrently |

**Overview of a Memory System**
Artificial human memory has been modeled as employing cues in a routine to find matches within associative memory (Burger 2009).  One may envision an anthropomorphic system of dynamic (short term) working memory as in Fig. 1.  Working memory has been modeled as orchestrating long term memory (Burger Aug 30, 2010).  Long term memory in humans is analogous to PROM (programmable read only memory that may be programmed exactly once).  In humans, the contents of working memory are known (via consciousness) but all else, including editing processes are hidden.

An important albeit neglected attribute of any memory system is its ability to "learn."  Learning is definitely not memorization.  Learning here means an ability to run a sequence of memory words without controls from working memory (or CPU).  This ability occurs only if the sequence is sufficiently rehearsed and only if a word of the sequence has been addressed by working memory.  This type of learning has been termed "state machine" learning.

**Artificial Leaning for Artificial Memories**
A word of long term memory is assumed to hold components for images or actions.   Each word in a practical system can be diagramed as in Fig. 2.  The signals and commands held by the word are released by activating the *enable*.  Once the signals and commands have all been discharged, there is a *done* signal.

To learn something that is being practiced, a *timing filter* is required to detect a repeated sequence. For example the filter must detect whenever a certain word is enabled immediately after another given word is done. A suitable digital filter to detect a sequence exploits the delay between active words, specified to be *Delay 1* as in Fig. 3. Part (a) of the figure indicates that a *done* signal from Word 1 is held for a time *Delay 1*. Implementation of Delay 1 is not shown but is straightforward using elements of short term memory. When a second *done* signal arrives in a timely way from some other word, say from Word 2, the AND gate is activated to make $X_1$ true.

$X_1$ results in the generation of a single spike. The spike operates a shift register of set-only D-latches identified as $D_1, D_2 \ldots D_{n-1}, D_n$. All D-latches initially are set to Boolean zero (cleared). Each time the 1-2 sequence occurs, a true signal shifts to the left one place in the shift register.

Latch details appear in part (b) of the Fig. 3. Note that the D-latches can be set true only once in this plan, since learning, once it occurs, is assumed long term. The spike generator in part (c) of the figure uses a standard XOR (exclusive OR) to produce a brief spike on the leading edge of signal $X_1$. Spike width is determined by *DELAY 2*. Spike width is just enough to move the Boolean signal $X_1$ only one step along the shift register. AND, OR and XOR gates are readily available both in brain neurons and in hardware (Burger Apr 2010).

The design in Fig. 4 supports possible state transitions from any given word to any other given word in a block of n words within long term memory. The maximum number of possible sequences is n! Fig. 4 shows only three words, although many more can be used. In an attempt to gain focus, consider the sequence 1-3-2 that has been used, say, ten times so that it is well learned. Then, whenever Word 1 is addressed (associatively) by short term memory a minor miracle occurs. After a time equal to *Delay 1*, Word 3 is addressed directly by Word 1. Then after another time equal to *Delay 1*, Word 2 is addressed directly by Word 3. This will accomplish a three-step procedure without CPU involvement.

Note that switches $S_{ij}$ ($1 \leq i \leq K$, $1 \leq j \leq K$, $i \neq j$) enable the state machine. For example, a switch $S_{13}$ applies the done signal from Word 1 to the delayed enable of Word 3. Switches are viewed by neuroscientists as synaptic connections promoted by interneurons, but computer engineers create contacts artificially with a FET (Field Effect Transistor).

In general for K words and an arbitrary sequence that may go forward or backward, K (K-1) timing filters are needed. This implies a like number of bus lines for the filter outputs and a like number of switches $S_{ij}$ ($1 \leq i \leq K$, $1 \leq j \leq K$, $i \neq j$). Also there must be K lines for memory word outputs and K lines for memory word inputs. The overhead is reasonable for systems with complexity such as humanoid memory.

**Limitations of Artificial Learning**
A pitfall is that once a learned action is activated, it executes asynchronously. In practical memory, as in human memory, it is always possible to override learning. In this case, one simply opens the $S_{ij}$ circuit using another series switch (not shown). This returns control to working memory same as it does in humans.

Learning cannot be erased easily, so if the learning is wrong, a correct version must be re-learned. This is also true of human learning.

A limitation of any state machine is that states (words) must be distinguishable. This means that a learned sequence cannot repeat a given word in a sequence. For example, one cannot learn the sequence 1-3-1-2 because state 1 is repeated. However, different versions of a repeated word are readily possible, permitting for example, 1a-2-1b-3. Looping is not automatic since it would involve repeating a word, for example, 1-2-3-1-2-3…. In this system a given sequence may be repeated by only a call from working memory.

**Conclusions**
Learning is largely important both to humans and to intelligent robots. Survival matters, which hinges on reflexive actions concurrent with quick thinking, especially when there is no time to ponder the pros and cons of a response. From another view, learning enables multitasking in the sense that while learned actions are being performed one may think about something else. One may read music and also play the violin, for instance.

Learning as above depends on timing filters that anyone can build and verify. These filters detect and remember a sequence that is rehearsed a given number of times (n). Learned sequences, once called, run automatically within long term memory. CPU effort is not required. Obviously, this frees central processing for more important work while a sequence is running. More importantly, long term memory is freed from the constraints of CPU control. Consequently artificial learning creates the possibility of parallel processing since many learned sequences may run simultaneously assuming they involve independent memory words.

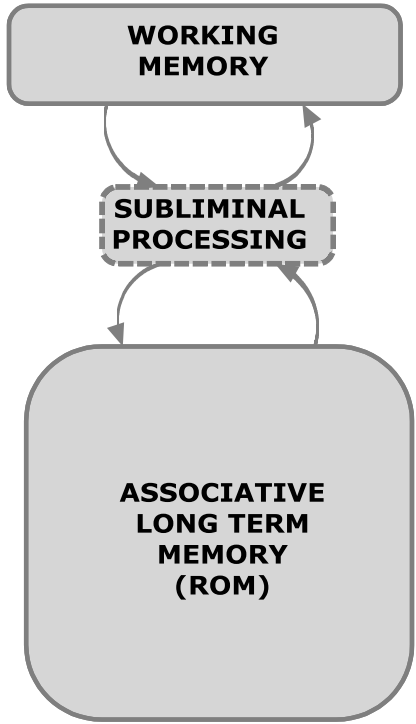

Fig. 1  Short term (working) memory (consciousness) provides cues that are edited subliminally while recalls from associative long term memory are tested unconsciously before refreshing working memory

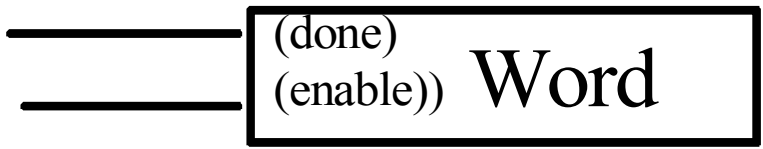

Fig. 2  Symbol for a word of long term memory with controls

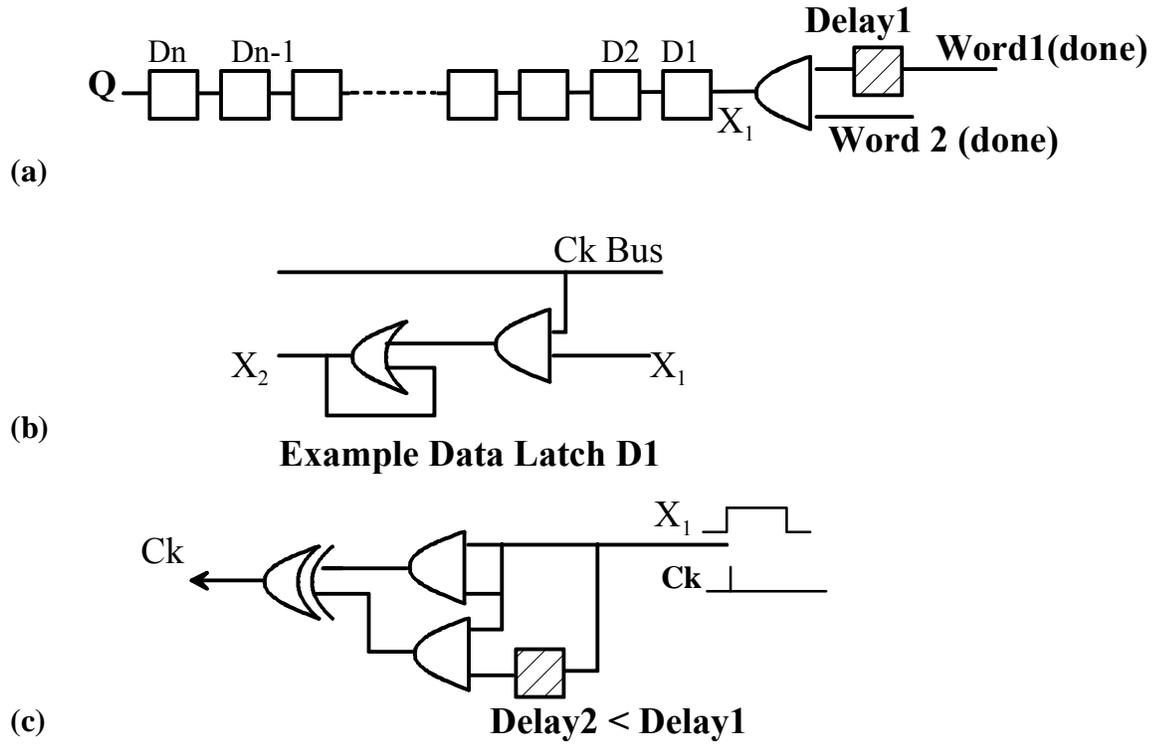

Fig. 3  Learning filter showing how rehearsal technically results in permanent learning

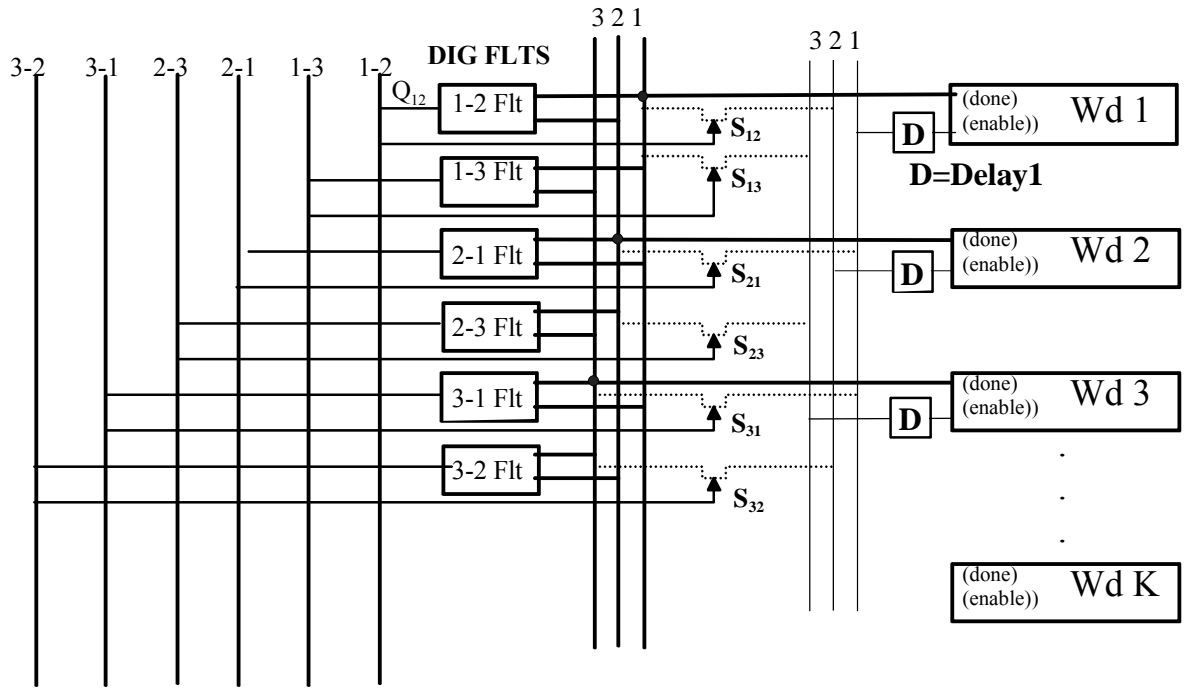

**Fig. 4** Embedded state machine to learn an arbitrary sequence of words